\documentclass{article}
\usepackage{spconf,amsmath,graphicx}

\title{Improving Generalization of Transformer for Speech Recognition with Parallel Scheduled Sampling and Relative Positional Embedding}
\name{Pan Zhou$^{1}$\sthanks{\hspace{1pt} This work was done as a postdoctor at Tsinghua University before joining Sogou.}, Ruchao Fan$^{2}$\sthanks{\hspace{1pt} Participated in this work as an intern at Sogou before entering UCLA.}, Wei Chen$^{1}$, Jia Jia$^{3}$
}
\address{$^{1}$AI Interaction Division, Sogou Inc., Beijing, P.R.China
\\ $^{2}$Department of Electrical and Computer Engineering, University of California, Los Angeles, USA
\\ $^{3}$Department of Computer Science and Technology, Tsinghua University, Beijing, P.R.China
\\\small \texttt{\{zhoupan,chenweibj8871\}@sogou-inc.com,} \texttt{fanruchao@g.ucla.edu,} 
\texttt{jjia@mail.tsinghua.edu.cn}
}

\begin{document}
%
\maketitle
\begin{abstract}
Transformer has shown promising results in many sequence to sequence transformation tasks recently. It utilizes a number of feed-forward self-attention layers 
to replace the recurrent neural networks (RNN) in attention-based encoder decoder (AED) architecture. Self-attention layer learns temporal dependence by incorporating sinusoidal positional embedding of tokens in a sequence for parallel computing. Quicker iteration speed in training than sequential operation of RNN can be obtained. Deeper layers of the transformer also make it perform better than RNN-based AED. However, this parallelization ability is lost when applying scheduled sampling training. Self-attention with sinusoidal positional embedding may cause performance degradations for longer sequences that have similar acoustic or semantic information at different positions as well. To address these problems, we propose to use parallel scheduled sampling (PSS) and relative positional embedding (RPE) to help the transformer generalize to unseen data.
Our proposed methods achieve a 7\% relative improvement for short utterances and a 70\% relative gain for long utterances on a 10,000-hour Mandarin ASR task.

\end{abstract}

%
\section{Introduction}
\label{sec:intro}

End-to-end (E2E) ASR aims to simplify conventional ASR by jointly learning acoustic model (AM), pronunciation model (PM) and language model (LM) within one single neural network and has achieved promising results. Connectionist Temporal Classification (CTC) \cite{graves2006connectionist,amodei2016deep}, Recurrent Neural Network Transducer (RNN-T) \cite{graves2012sequence,battenberg2017exploring,rao2017exploring}, Recurrent Neural Aligner \cite{sak2017recurrent,dong2018extending}, Segment Neural Transduction \cite{yu2016online} and Attention-based encoder decoder (AED) models \cite{chorowski2014end, chan2016listen, chiu2018state} are such E2E models that are well explored in the literature. 

AED model \cite{chorowski2014end,chan2016listen} was examined on many speech tasks.
AED consists of an encoder, a decoder and an attender which bridges encoder and decoder.
The encoder often uses RNNs to capture higher level temporal features from raw input spectral features. The decoder generates tokens sequentially conditioned on the context tokens and the output of attender. While ground truth labels are used as context tokens during training, the context tokens for decoder input are the predicted tokens when doing inference. Scheduled sampling (SS) \cite{bengio2015scheduled} is often adopted to compensate the discrepancy between training and inference.
Step-by-step predictions of RNN are also suitable for SS to be integrated at each token level without sacrificing training efficiency. However, this sequential feature of RNN makes the training time-consuming.

An alternative to RNNs in AED is self-attention layer which was proposed in Transformer \cite{vaswani2017attention} for neural machine translation (NMT). The self-attention architecture learns temporal and contextual dependency inside the input sequence by employing temporal attention on the input feature itself. Transformer has been applied to E2E speech recognition systems \cite{dong2018speech, zhou2018syllable,zhou2018comparison,karita2019comparative} and has achieved promising results. The transformer-based E2E ASR relies on feed-forward self-attention components and can be trained faster with more parallelization than RNN based AED.
However, the non-recurrent parallel forward process of decoder in transformer also makes it inefficient to utilize scheduled sampling at training stage. Sequence order of speech, which can be represented by recurrent processing of input features, is an important distinction. Although absolute positional embedding (APE) added to input features can make the transformer be aware of the order in a sequence, performance degradation of long sentences of AED, as mentioned in \cite{chorowski2015attention}, may become more serious for transformer. Compared to RNN-based AED, we indeed found that vanilla transformer is more sensitive to sentences that are longer than those in the training set. 




In this paper, we propose to use parallel scheduled sampling (PSS) for efficient SS in transformer. 
More precisely, we mixed the ground truth label with the output of Kaldi-based hybrid model or transformer itself to form a new input to the decoder, so as to simulate the error distribution of transformer during inference and maintain the parallelization of decoder.
Experimental results show that we could get
about 7\% relative gain compared to teacher-forcing training. 

Transformer also makes lots of deletion errors for long utterances. Characters whose positions exceed the max length of training data are likely to be deleted. We name this deletion as tail deletion (TD).
Other deletions appear between similar acoustic or semantic segments and we call such deletions as internal deletion (ID).
We argue that the attention mechanism accessing the whole sentence makes the model confused of focusing on unseen long sentences.
In order to restrict the attention position range to avoid model from this confusion, relative position embedding is proposed to help the model generalize to unseen longer sentences and 30\% absolute gain is obtained on test set with utterances longer than training set.

The rest of paper is organized as follows. Section \ref{sec:review_transf} briefly reviews transformer used in ASR tasks.  The proposed methods are described in details in Section \ref{sec:proposed_methods}. Experimental setup and results are presented in Section \ref{sec:exp} and the paper is concluded with our findings and future work in Section \ref{sec:conclusions}.

\section{Transformer-based E2E ASR}
\label{sec:review_transf}

The AED used in ASR is common to other sequence to sequence task.
The encoder transforms an input sequence of spectral features $(\boldsymbol{x}_1,..., \boldsymbol{x}_n)$ to a higher level representation. Conditioned on the high level representation, the decoder predicts output sequence $(y_1,..., y_u)$ of speech modeling units, such as phones, syllables, characters or sub-words (BPE \cite{kunchukuttan2016learning}), auto-regressively. 
Transformer follows AED architecture and uses stacked multi-head attention (MHA) and point-wise fully connected feed-forward network (FFN) for both encoder and decoder \cite{vaswani2017attention}.



\subsection{Multi-head Attention}
\label{ssec:dot_att}
The attention layer used in transformer uses the "scaled dot-product attention" with the following form:
\begin{equation}
Att(Q,K,V)=softmax(\frac{QK^T}{\sqrt{d_k}})V,
\label{eq:scale_dot_att}
\end{equation}
where $Q$ and $K$ are queries and keys of dimension $d_k$, $V$ are values of dimension $d_v$. In order to allow the model to pay attention to different representation subspace, \cite{vaswani2017attention} proposed to use multi-head attention to perform parallel attention:
\begin{equation}
\label{eq:mha_1}
MHA(Q,K,V)=Concat[H_1,H_2,...,H_h]W^O,
\end{equation}
\begin{equation}
\label{eq:mha_2}
H_i=Att(QW_i^Q,KW_i^K,VW_i^V),
\end{equation}
where $W_i^Q \in R^{d_m \times d_k}$,$W_i^K \in R^{d_m \times d_k}$,$W_i^V \in R^{d_m \times d_v}$ and $W^O \in R^{hd_v \times d_m}$ are learnable weight matrices, $h$ is the total number of attention heads, $H_i$ is the output of the $i$-th attention head, $d_k$ is the individual dimension for each attention head, $d_m$ is the model feature dimension. 

For encoder and self-attention layers in decoder, all of the keys, values and queries, $K,V,Q$,  come form the output features of previous layer. In the source attention layers, queries come from the previous decoder layer, and the keys and values come from the final output of encoder. 

\subsection{Absolute Positional Embedding}
\label{ssec:pe}
Transformer contains no recurrence and convolution. In order to make use of the order of the sequence, input representations are added with absolute positional encoding before feeding to the encoder and decoder stacks.
\begin{equation}
\label{eq:sin_pe} 
PE_{(pos,i)}=
\left\{
\begin{aligned}
sin(pos/10000^{i/d_m}), \ {i\ is\ even,} \\
cos(pos/10000^{(i-1)/d_m}), \ {i\ is\ odd.}
\end{aligned}
\right.
\end{equation}
where $pos$ represents the absolute position in sequence, $i$ represents the $i$-th dimension of input feature.
For more details please refer to \cite{vaswani2017attention,dong2018speech}.


\section{Proposed Methods for Improving Generalization of Transformer}
\label{sec:proposed_methods}

\subsection{Parallel Scheduled Sampling}
\label{ssec:pss}
Scheduled sampling is a training strategy to help the auto-regressive models be robust to prediction errors during inference. However, vanilla transformer, when applied with scheduled sampling, is slower than teacher-forcing training
due to the destructed parallelism and duplicated calculations. In order to alleviate the problem, we implement a parallel scheduled sampling (PSS) mechanism, which obtains all the input tokens of decoder in advance by simulating the error distribution of inference during training. The PSS mechanism is very compatible with the training of transformer. Specifically, two error distribution simulation methods for PSS are proposed. Unlike scheduled sampling in \cite{li2019speechtransformer}, our methods consider the sampling at each decoding step without losing efficiency.

\subsubsection{PSS with simulation from chain model}
\label{sssec:chain pss}
In this method, the error distribution of transformer outputs is simulated with the decoding results of the Kaldi-based chain model (CM) \cite{povey2016purely}. For each utterance, we first obtain its hypothesized text $\boldsymbol{\hat{y}}=(\hat{y_1},..., \hat{y_q})$ from a previously trained chain model, as the simulated outputs of transformer. Then, $\boldsymbol{\hat{y}}$ is mixed with ground truth $\boldsymbol{y}=(y_1,..., y_u)$ according to a teacher-force rate to get the final decoder input $\boldsymbol{\overline{y}}=(\overline{y_1},..., \overline{y_u})$. The teacher-force rate represents the probability of choosing the tokens in the true label $\boldsymbol{y}$ as the decoder input and varies according to the schedule with the following piece-wise linear function:
\begin{equation}
\label{eq:chain pss}
P(i) = max(min(1, 1 - (1-P_{min})*\frac{i - N_{st}}{N_{ed}-N_{st}}), P_{min})
\end{equation}
where $P_{min}$ is the minimum teacher-force probability which is usually not zero to prevent under-fitting over training set, $N_{st}$ and $N_{ed}$ represent the starting and the ending step in the schedule respectively, and $i$ is the training step. The step $i$ in Eq. \ref{eq:chain pss} can be either $epoch$ or $batch$.

After obtaining the teacher-force rate, $sentence$ or $token$ level token mixing methods can be considered. The $sentence$ level mixing is more explicit. It is achieved by selecting either $\boldsymbol{\hat{y}}$ or $\boldsymbol{y}$ as $\boldsymbol{\overline{y}}$ according to the teacher-force rate. On the contrary, the $token$ level is closer to the real scheduled sampling by considering mixing of the token at each step. The process can be formulated as:
\begin{equation}
\label{eq:by token}
\overline{y_j} = \left\{
\begin{array}{ccl}
y_j & & {with\ P(i)} \\
\hat{y_j} & & {with\ 1-P(i)\ and\ j \leq q} \\
pad & & {with\ 1-P(i)\ and\ j > q}
\end{array}
\right.
\end{equation} 
where $j=1,2,\cdots,u$ and $pad$ is the padding token to fill up short sentences in a batch. The token sequences $y$ in this paper are sequences of scalar. At last, the mixing decoder input $\boldsymbol{\overline{y}}$ is fed to the transformer at the same time to achieve parallelization.
Although there exists an increased time cost of obtaining the hybrid system results of each utterance, it is much faster for PSS training than the training of the transformer with traditional scheduled sampling.

Our method is related to \cite{sennrich2015improving-backNMT} which adopts back-translation to get source text with errors to be used for training.  Scheduled sampling aims to make AED robust to several decoding errors. Although there is no evidence that the error distributions of hybrid system have any relation to that of AED, hybrid system results, containing several errors, can be viewed as a special kind of data augmentation technique. So it should help the model to generalize.

\subsubsection{PSS with simulation from self-decoding}
\label{sssec:self pss}
The second method we describe here is to simulate the error distribution by the results decoding from transformer itself. The intuitive way is to generate hypothesized text $\boldsymbol{\hat{y}}$ for train set in advance with teacher-forcing-trained model. Furthermore, online training mechanism is applied in our experiments, in which the $\boldsymbol{\hat{y}}$ are generated batch by batch with the newest updated model. To simplify the online training process, the hypothesized results are obtained by greedy search decoding with ground truth $\boldsymbol{y}$ as decoder input. Since the only difference between this method and the method in Section \ref{sssec:chain pss} is how we simulate the error distribution of transformer output, the same scheduled strategies as in Eq. (\ref{eq:chain pss}) is used.

Recently, we find that the method in this part is a special case in \cite{duckworth2019parallel}, which is proposed for machine translation. They also proposed a parallel scheduled sampling mechanism to achieve parallelism in transformer, where they consider $N$ times of token mixing. Hence, we reveal that teacher forcing training is the special case when $N=0$ and our method is the case when $N=1$. To associate them with conventional scheduled sampling, we can make the following reasoning process. When $N=1$, only the first generated character $\overline{y_1}$ is sampled from decoding. If $\overline{y_1}$ is fed to decoder and repeat the above operation, both $\overline{y_1}$ and $\overline{y_2}$ are sampled reasonably. Therefore, repeating above process for $N>=u$ times, PSS with self-decoding result is equivalent to the conventional scheduled sampling. The time cost will also be approximately the same as conventional one.

\subsection{Relative Positional Embedding}
\label{ssec:rpe}
We observe many deletion errors for speech transformer, such as TD and ID in long utterances. Many works attempted to solve the long sequence generation problem by increasing training data length. In this direction, their main work focused on memory controlling by sparse attention mechanism or segmental transformer \cite{child2019generating,huang2018music,dai2019transformer}. \cite{narayanan2019recLongSpeech} proposed to use diverse acoustic data and LSTM state manipulation to simulate long audio for training to improve generalization on unseen long speech.
Using longer training sentences is indeed a solution to the TD problem but the ID problem may not be solved because of confusion among similar segments in the global self-attention. How to make model generalize well when there is only short data is our focus.

The original APE proposed for NMT may not be suitable for speech recognition applications, since alignment between text and speech is monotonic.
APE may make model attend to wrong positions for long utterance with similar segments since it is never trained on such long sentences. Moreover the attention scores only focus on a local area for speech recognition. Thus we want to use relative positional embedding for vanilla speech transformer to try to explore how the positional embedding affects long sentence recognition performance.
The idea of relative positional encodings
has been previously explored in the context of machine
translation \cite{shaw2018self}, music generation \cite{huang2018music} and language modelling \cite{dai2019transformer}. Recently, transformer transducer with RPE \cite{zhang2020transformer} is proposed for streaming speech recognition. The main different between ours and \cite{zhang2020transformer} is that we explore RPE in the attention based encoder decoder framework other than RNN transducer framework and we systematically evaluate its impact on speech recognition especially for the model collapse problem on unseen long data mentioned above.

Our main purpose is to restrict the position range and strengthen the relative position relationship within the range. 
The TD may be alleviated by RPE by letting the model learn how to pay attention to relative context. Furthermore, fewer similar segments appear in restricted position range, resulting in solving the ID problem.
Suppose the input of a self-attention layer is $\boldsymbol{z}=(\boldsymbol{z}_1,..., \boldsymbol{z}_T)$ which is a sequence of vector, we define the relative position between each $\boldsymbol{z}_i$ and $\boldsymbol{z}_j$ as $a_{ij}$. To restrict the position range, we consider a maximum absolute value of relative position $k$ and acquire $2 \times k+1$ embeddings, denoted as $w = (w_{-k},\cdots,w_k)$. Then, any relative positional embedding between two inputs can be formulated as:   
\begin{equation}
a_{ij} = w_{max(-k,min(k, j-i))}
\end{equation}
Next, the relative position embedding $a_{ij}$ is incorporated to the similarity computation (softmax input) in Eq. (\ref{eq:scale_dot_att}) as follows:
\begin{equation}
\label{eq:revised energy}
e_{ij} = \frac{z_iW^Q(z_jW^K+a_{ij})^T}{\sqrt{d_k}}
\end{equation}
where the notations are consistent with the equations in Section \ref{sec:review_transf}.
Eq. (\ref{eq:revised energy}) can be modified by applying distributional law of matrix split into two terms for efficient training:
\begin{equation}
e_{ij} = \frac{z_iW^Q(z_jW^K)^T+z_iW^Q(a_{ij})^T}{\sqrt{d_k}}
\end{equation}

We use different RPE across layers and share RPE among attention heads within the same layer. The first term is equivalent to the original similarity computation. 
The second term can be calculated with tensor reshaping, which means a matrix with size $bhn \times d_k$ multiply a matrix with size $d_k \times n$. And then it is reshaped to fit with the first term. $b$ is the batch size, $n$ is the sequence length.

\section{Experiment}
\label{sec:exp}

Our experiments are conducted on a 10,000 hour Chinese speech dictation data. We discard sentence longer than 40 Chinese characters. The main test sets include $\sim$ 33K short utterances (SU) and $\sim $5.6K long utterances (LU). The length distribution is listed in Table \ref{tableLength_data}.

71 dimension fbanks are extracted every 10 ms within a 25-ms window using the conventional ASR front end. Every four consecutive frames are stacked to form a 284-dimensional feature vector and we jump 4 frames to get shorter input feature sequence for transformer models.
\begin{table}[htb]
\centering
\caption{\textit{Sentence number of different length in training and test sets.}}
\begin{tabular}{|c||c|c|c|c|c|}
\hline
  \#char. & $\le$ 10 & (11,40)&(41,80)&$\ge$ 80    \\ \hline
training	&6,996,966 & 4,807,048 & 974& 0 \\ \hline
SU	&26863	& 6710&0 &0 \\ \hline 
LU  &  0 & 1 & 5407&248 \\ \hline
\end{tabular}
\label{tableLength_data}
\end{table}

6812 Chinese characters are used as modeling unit. 
All models are trained by optimizing the cross entropy 
via adam optimizer. Label smoothing (LS) is set to 0.1 during training to improve performance. We also apply 10\% dropout rate to the output of each sub-layer, before it is added to the sub-layer input and normalized. The total training epoch is fixed to 12. We train our model from random initialization with an initial learning rate of 0.0002 and halve it from epoch 7. Beam search with beam width of 5 is used without external language models to evaluate our model on test sets.

All our transformers contain 5 MHA-FFN blocks and 3 MHA-MHA-FFN blocks with 16 attention heads. The size of FFN layer is 2048 and $d_m = 768$. A two layer FFN with 2048 and 768 nodes is placed before the first encoder block. With the same 10,000-hour training data, We also train an LAS and a streaming TDNN-LSTM chain model for comparison, which are denoted as B0 and B2 respectively. The LAS is constituted of 4-layer BLSTM encoder and 1 LSTM decoder with the same base settings as in \cite{onlineLAS}. The CER of our base models are presented in the first block of Table \ref{tableCER_PSS}. Comparing with $B0$ and $B1$, we find that transformer performs better than LAS on SU but degrades a lot on LU.


\begin{table}[htb]
\centering
\caption{\textit{Performance in CER [\%] of PSS for transformer. $P_{min}$ is teacher force rate and $N$ is number of decoding pass.}}
\begin{tabular}{|c|c||c|c|c|c|}
\hline
 ExpID& Model  &$P_{min}$  &N & 	SU & LU   \\ \hline
B0	&	LAS 	& 	0.8  &	- &   10.78    &  23.32  \\ \hline
B1	&transformer&1.0	& -  	& 9.57  &   42.41   \\ \hline
B2  &  CM   &-&- &10.52 & 8.00 \\ \hline \hline 
E1&PSS-CM	&	0.7 	&-&8.9 &41.28 \\ \cline{3-6}
	&		&	0.8 	&-&8.93 &41.19 \\ \hline
E2& PSS-offline	&	0.8	&- &9.00 & 40.00 \\ \hline \hline
E3& PSS-online	&	0.5 	&1&8.88&42.47 \\ \cline{3-6}
	 &		&	0.5 	&2&8.93&41.52 \\ \hline
\end{tabular}
\label{tableCER_PSS}
\end{table}

\begin{figure*}[htb]
\centerline{\includegraphics[width=0.94\linewidth]{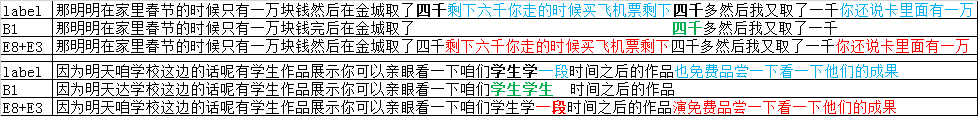}}
\caption{Decoding examples, showing tail deletions and internal deletions of $B1$, on LU test set for various systems.}
\label{fig:example}
\end{figure*}

\subsection{Improvements by PSS}
\label{ssec:exp_pss}
Due to lack of space, we only report experiments with $token$ and $batch$ style as mentioned in section \ref{sssec:chain pss}. 
Using the model in $B1$ as a starting point, we perform several PSS experiments. First, we perform PSS with decoded text results from TDNN-LSTM chain model trained with 50,000 hours of data, numbered as $E1$. Since CM used here is trained with more data than $B2$, it performance better on SU and LU, which is 6.65\% and 5.32\% respectively.
With $P_{min}=0.8$, we obtain a 8.93\% of CER on SU. One may argue that this gain may come from the larger training data of the CM that provides decoding results for transformer. So we conduct $E2$ with hypothesized text of training data generated by beam search decoding with transformer of $B1$. $E2$ achieves a similar result with $E1$. In $E3$, we generate hypothesized text during training stage in an online fashion to make it closer to real scheduled sampling process. We reach a CER of 8.88\% on SU with $N=1$ and $P_{min}=0.5$, about a 7.2\% relative reduction.
Better results achieved by $E3$ show that decoding as training proceeds is a better choice for parallel scheduled sampling.

\subsection{Improvements by RPE}
\label{ssec:exp_rpr}

As mentioned above, LAS generalizes better on LU than transformer because of its iterative process of sequence, which learns a better order information. 
It reveals that order information learned by self-attention layers with APE generalize poorly to sequences that are not seen in the training set. 


Experimental results are summarized in Table \ref{tableCER_RPR}. We first investigate APE in encoder. In $E5$, we remove APE in encoder and keep APE in decoder. Although SU drops a lot, LU even improves, which indicates APE can not generalize well on long sentences. In $E6$, we replace the fixed APE with a learnable token ID APE in encoder, just like the token embedding in the decoder. Token ID embedding is similar on SU but worse on LU than sin/cos, which may because token ID can not be learned for positions not seen in data and sin/cos is fixed in advance. Then we try to introduce RPE in encoder ($E7$) and decoder ($E8$) gradually. When RPE is used in each MHA layer besides adding APE to input feature of encoder block, LU improves from 42.41\% to 33.56\%. If we discard APE and set $k=10$ in encoder, it is further improved to 29.87\%. $k=10$ represents attending to about 2 Chinese characters length (40 frames) on both sides of current query of self-attention in encoder. As $k=10$ performs better than $k=40$, we guess it is enough for acoustic self-attention representation learning in transformer encoder. The CER continues to decrease to 12.73\% as we utilize a 2-character-range RPE in the decoder MHA to replace APE. With RPE, although such long sentences have not been presented to the model, the local attention relationship has already been learned. Besides, the possibility for similar segments to appear in the near context becomes smaller, which decreases the burden of model to distinguish the similar segments. Thus, RPE helps to decrease TD and ID. This also indicates that local and relative position is more suitable for speech recognition. Overall, with relative position embedding, we lower the CER of LU set from 42.41\% to 12.73\%, an absolute 30\% gain.

\begin{table}[htb]
\centering
\caption{\textit{Effects in CER [\%] of APE and RPE for transformer. 10-4 represents position range is 10 for encoder and 4 for decoder. $k$ is the relative position range in RPE.}}
\begin{tabular}{|c||c|c|c|c|c|}
\hline
  ExpID 	& Enc 		&Dec&   $k$      & SU & LU   \\ \hline
B1    		& APE  		& APE	&	-&9.57    &  42.41  \\ \hline \hline
E5		& -			& APE &	-&10.31	&  37  \\ \hline 
E6		& token ID APE	& APE & 	-&9.58	&47.65 \\ \hline \hline
E7		& APE +RPE		&   	&20	&9.17 &33.56 \\ \cline{4-6}
		& RPE			& 	& 40 &9.38&32.22 \\ \cline{4-6}
		&RPE			&APE	&20 & 9.17&31.17 \\ \cline{4-6}
		&RPE		&   	& 10&9.24 &29.87 \\ \hline \hline

E8		&RPE		&RPE	&10-2	&9.31&12.73 \\ \cline{4-6}
		& 	&           &10-4	&9.2&12.89 \\ \hline \hline
E8+E3		&   RPE    &RPE  & 10-4 & $\boldsymbol{8.9}$ & $\boldsymbol{12.89}$ \\ \hline
\end{tabular}
\label{tableCER_RPR}
\end{table}


\subsection{Combination of PSS and RPE}
\label{ssec:exp_final}

Then, $E8+E3$ in last row of Table \ref{tableCER_RPR} shows the results of combination of RPE and PSS with settings in first row of $E3$. SU continues to be improved to 8.9\% and LU stays unchanged. The mixing process with $N=1$ may be the reason of none improvement on LU.
Compared to the base transformer $B1$, there is a 7.0\% relative improvement of CER on SU and a 70\% relative improvement on LU.

\begin{figure}[htb]
\centerline{\includegraphics[width=0.9\linewidth]{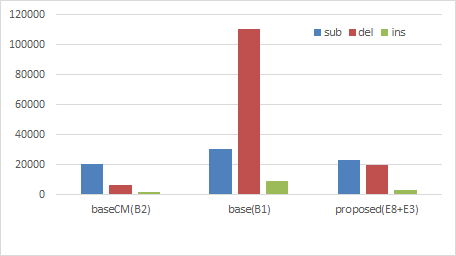}}
\caption{Error number comparison on the LU test set for various systems.}
\label{fig:errors}
\end{figure}

Error analysis on LU set is shown in Fig \ref{fig:errors}, from which we observe that deletions drop significantly from $B1$ to $E8+E3$. Lastly, we examine some decoding examples
in Fig. \ref{fig:example}. The length of $B1$ output is about 40 characters and characters after 40 is deleted. Attention jump from the first "four thousands" to the second "four thousands" in the first example can also be observed. Tail deletions and internal deletions of vanilla transformer in $B1$, blue characters in labels, are recognised by RPE transformer in $E8+E3$ which are shown in red. 
Also, RPE alleviates the self-loop problem in AED decoding as shown in green in the second example.


\section{Conclusions and Future Work}
\label{sec:conclusions}

In this work, we apply PSS and RPE to successfully improve the generalization ability of transformer. PSS simulates the real sampling process of SS efficiently and alleviate the discrepancy betweent training and testing. RPE endues transformer the ability to distinguish similar acoustic and semantic segments and ease the model collapse problem for unseen long sentences.
Experiments show 7\% relative improvement for short utterances and a 70\% relative gain 
for long utterance is achieved.
To further improve the generalization of transformer in speech, attention restriction in source attention is our next direction.

\vfill\pagebreak



\bibliographystyle{IEEEbib}
\bibliography{refs}

\end{document}